\documentclass[10pt,twocolumn,letterpaper]{article}

\usepackage[pagenumbers]{cvpr}
\usepackage{amsmath}
\usepackage{amssymb}
\usepackage{booktabs}
\usepackage{amsfonts}
\usepackage{algorithmic}
\usepackage{algorithm}
\usepackage{array}
\usepackage{textcomp}
\usepackage{stfloats}
\usepackage{url}
\usepackage{verbatim}
\usepackage{graphicx}
\usepackage{cite}
\usepackage{xspace}
\usepackage{threeparttable}
\usepackage{xcolor}
\usepackage{multirow}
\usepackage[T1]{fontenc}
\usepackage{gensymb}
\usepackage{booktabs}
\usepackage{amssymb}
\usepackage{enumitem}
\usepackage[hyphenbreaks]{breakurl}
\usepackage{mathtools}
\usepackage{xspace}
\usepackage{diagbox}
\usepackage{graphbox}
\usepackage{bm}

\makeatletter
\newcommand*{\rom}[1]{\expandafter\@slowromancap\romannumeral #1@}

\makeatother

\newcommand{\FLIP}{\protect\reflectbox{F}LIP\xspace}

\usepackage[pagebackref,breaklinks,colorlinks]{hyperref}

\usepackage[capitalize]{cleveref}
\crefname{section}{Sec.}{Secs.}
\Crefname{section}{Section}{Sections}
\Crefname{table}{Table}{Tables}
\crefname{table}{Tab.}{Tabs.}

\def\cvprPaperID{12650} 
\def\confName{CVPR}
\def\confYear{2023}

\begin{document}

\title{Learning a Deep Color Difference Metric for Photographic Images}

\author{Haoyu Chen\textsuperscript{\rm 1}\quad Zhihua Wang\textsuperscript{\rm 1,2,}\thanks{Corresponding author.}\quad Yang Yang\textsuperscript{\rm 3}\quad Qilin Sun\textsuperscript{\rm 4}\quad Kede Ma\textsuperscript{\rm 1}\\
\normalsize \textsuperscript{\rm 1}City University of Hong Kong \quad
\normalsize \textsuperscript{\rm 2}Shenzhen MSU-BIT University \\
\normalsize \textsuperscript{\rm 3}Shenzhen Transsion Holdings Co., Ltd.  \quad
\normalsize \textsuperscript{\rm 4}The Chinese University of Hong Kong (Shenzhen)\\
{\tt\small  \{haoychen3-c,zhihua.wang\}@my.cityu.edu.hk, 
 kede.ma@cityu.edu.hk,}\\
{\tt\small yang.yang6@transsion.com, sunqilin@cuhk.edu.cn}}
\maketitle

\begin{abstract}
Most well-established and widely used color difference (CD) metrics are handcrafted and subject-calibrated against uniformly colored patches, which do not generalize well to photographic images characterized by natural scene complexities. Constructing CD formulae for photographic images is still an active research topic in imaging/illumination, vision science, and color science communities. In this paper, we aim to learn a deep CD metric for photographic images with four desirable properties. First, it well aligns with the observations in vision science that color and form are linked inextricably in visual cortical processing. Second, it is a proper metric in the mathematical sense. Third, it computes accurate CDs between photographic images, differing mainly in color appearances. Fourth, it is robust to mild geometric distortions (\eg, translation or due to parallax), which are often present in photographic images of the same scene captured by different digital cameras.  We show that all four properties can be satisfied at once by learning a multi-scale autoregressive normalizing flow for feature transform, followed by the Euclidean distance which is linearly proportional to the human perceptual CD. Quantitative and qualitative experiments on the large-scale SPCD dataset demonstrate the promise  of the learned CD metric. Source code is available at \url{https://github.com/haoychen3/CD-Flow}.

\end{abstract}

\section{Introduction}
\label{sec:intro}
 
For a long time in vision science community, the modular and segregated view of cortical color processing predominated~\cite{shapley2011color}: the visual perception/processing of color-related quantities is separate from and in parallel with the perception/processing of form (\ie, object shape and structure), motion direction, and depth order in natural scenes. As a result, vision scientists preferred to investigate color perception under minimal conditions on form~\cite{katz1935world, shapley2011color}, for example, using uniformly colored patches.

The idea that color as a visual sensation can be analyzed separately had a profound impact on the development of computational formulae for color difference (CD) assessment. Till now, the most well-established and widely used CD metrics are primarily built upon  the three-dimensional \textit{spatially-isotropic} CIELAB coordinate system \cite{mahy1994cielab}, recommended by the International Commission on Illumination (abbreviated as CIE from its French name Commission Internationale de l’{\`E}clairage) in 1976. However, the uniformity\footnote{A system is perceptually uniform if a small perturbation to a component value is approximately equally perceptible across the range of that value~\cite{poynton1996technical}.} of the CIELAB space is not as ideal as intended~\cite{luo1999colour}, even for uniformly colored patches. Thus, more complex and parametric formulae are proposed to rectify different aspects of perceptual non-uniformity. Representative methods include JPC79 \cite{mcdonald1980jpc79}, CMC($l$:$c$)\footnote{$l$ and $c$ are two multiplicative parameters in the model to be fitted.} \cite{clarke1984cmc}, BFD($l$:$c$) \cite{luo1987bfd}, CIE94 \cite{mcdonald1995cie94}, and CIEDE2000 \cite{luo2001ciede2000}, in which the parameters are calibrated by fitting the human perceptual CD measurements of uniformly colored patches. A na\"{i}ve application of these metrics to photographic images is to compute the mean of the CDs between co-located pixels, which has been empirically shown to correlate poorly to human perception of CDs~\cite{jaramillo2019evaluation}.

Back to the vision science community, with more supporting evidence from psychophysical and perceptual studies~\cite{lennie1999color, shevell2008color,ben2004hue, abertazzi2011perception}, vision scientists have gradually come to agree on an alternative and more persuasive view of color perception: color and form (and motion) are inextricably interdependent as a unitary process of perceptual organization~\cite{shapley2011color,kanizsa1979organization}. Even the primary visual cortex (\ie, V1)
plays a significant role in color perception through two types of color-sensitive neurons: single-opponent and double-opponent cells. The single-opponent cells are sensitive to \textit{large areas} of color, while the double-opponent cells respond to \textit{color patterns}, \textit{textures}, and \textit{boundaries}~\cite{land1977retinex,shapley2011color}. At later stages, color is transformed to more complex and abstract features, which represent the integral properties of objects, and remain consistent against the changes of the environmental illumination~\cite{nakayama1992experiencing, hansen2006memory}.

Inspired by these scientific findings, researchers and engineers began to take spatial context (\ie, local surrounding regions) into account, when designing CD formulae. Representative strategies include low-pass spatial filtering~\cite{zhang1997scielab}, histograming~\cite{hong2006new}, patch-based comparison~\cite{wang2004image}, and texture-based segmentation~\cite{jaramillo2019evaluation}.
Most recently, Wang \etal \cite{wang2022cdnet} established the largest photographic image dataset, SPCD, for perceptual CD assessment. They further trained a lightweight deep neural network (DNN) for CD assessment of photographic images in a data-driven fashion, as a generalization of several existing CD metrics built on the CIE colorimetry. Nonetheless, the learned formula may not be a proper metric, due to reliance on the possibly surjective mapping for feature transform.

In this work, we further pursue the data-driven approach. We aim to learn a deep CD metric for photographic images with four desirable properties.
\begin{itemize}
    \setlength\itemsep{-0.05cm}
    \item It is conceptually inspired by color perception in the visual cortex. The design of our approach should respect the view that color and form interact inextricably through all stages of visual cortical processing.
    \item It is a proper metric that satisfies non-negativity, symmetry, identity of indiscernibles, and triangle inequality. Such design has been proven useful for perceptual optimization of image processing systems~\cite{ding2021comparison}.
    \item It is accurate in predicting the human perceptual CDs of photographic images, with good generalization to uniformly colored patches.
    \item It is robust to mild geometric distortions (\eg, translation and dilation), which are often present in photographic images of the same scene captured with different camera settings or along different lines of sight.
\end{itemize}
We show that all the four desirable properties can be satisfied  at once by learning a multi-scale autoregressive normalizing flow (a variant of RealNVP~\cite{dinh2017density} to be specific) for feature transform, followed by Euclidean distance measure in the transformed space. More specifically, we achieve the first property by the squeezing operation (also known as invertible downsampling) in the normalizing flow, which trades space size for channel dimension. The second property is a direct consequence of the bijectivity of the normalizing flow and the Euclidean distance measure. We achieve the third property by optimizing the model parameters to explain the human perceptual CDs in SPCD~\cite{wang2022cdnet}. We achieve the fourth property by enforcing the normalizing flow to be multi-scale and autoregressive, in which the features at a particular scale are conditioned on those at a higher (\ie, coarser) scale. By doing so, our metric automatically learns to preferentially rely on  coarse-scale feature representations with more built-in tolerance to geometric distortions for CD assessment.

We conduct extensive experiments on the large-scale SPCD dataset~\cite{wang2022cdnet}, and find that our proposed metric, termed as CD-Flow, outperforms $15$ CD formulae in assessing CDs of photographic images, produces competitive multi-scale local CD maps without any dense supervision, and is more robust to geometric distortions. Moreover, we empirically verify the perceptual uniformity of the learned color image representation from multiple aspects. 

\section{Related Work}
In this section, we first review CD formulae in a broader context, and then discuss normalizing flow-based models, which are core to the proposed CD metric.

\noindent\textbf{CD Formulae}.
The CD assessment is necessary for day-to-day color control, and is indispensable for color matching in color industries. Admittedly, CD formulae have accelerated the instrumental pass/fail devices for color judgments, but much still needs to be done for complete satisfaction. The scientific investigation of perceptual CDs can be dated at least back to Young and Helmholtz, who proposed and developed the trichromatic theory of color, which is the foundation of the metameric color matching experiment. CIELAB~\cite{mahy1994cielab} is one of the most successful CD metrics recommended by CIE in 1976, and has been widely adopted in industry for a long time. However, the CIELAB color space is not perceptually uniform~\cite{luo1986chromaticity}, which motivates the development of CIE94~\cite{mcdonald1995cie94} and CIEDE2000~\cite{luo2001ciede2000} through the introduction of application-specific parameters. Other CIELAB-based CD metrics include JPC79~\cite{mcdonald1980jpc79},  BFD($l$:$c$)~\cite{luo1987bfd}, and CMC($l$:$c$)~\cite{clarke1984cmc}. The introduced parameters are primarily calibrated using uniformly colored patches, digital or printed, which are statistically and semantically different from photographic images. Thus, the generalization of these metrics to photographic images is somewhat limited, especially when misalignment due to geometric distortions is present.

\begin{figure*}
  \centering
  \includegraphics[width=0.98\linewidth]{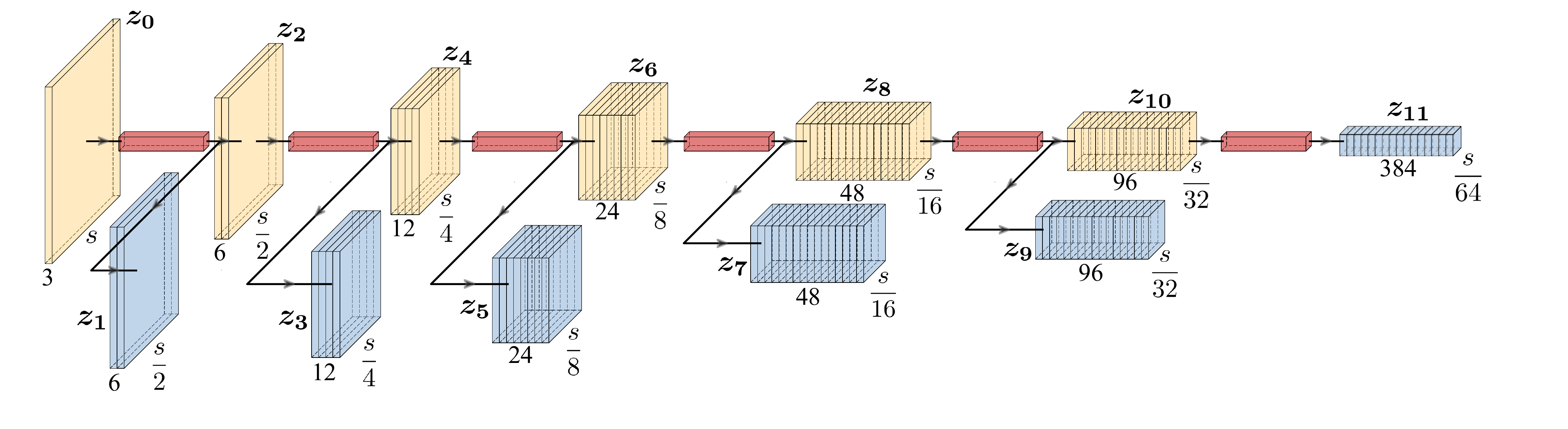}
  \caption{Feature transform of the proposed CD-Flow for perceptual CD assessment. The coordinate transform module consists of six scales. The leftmost yellow cube represents the input image with dimension $s\times s \times3$. The six blue cubes represent the six scales of representations, respectively. Each red cube represents a cascade of a squeezing operation, multiple flow steps, and a splitting operation, in which each flow step is composed of an actnorm operation, an invertible $1\times 1$ convolution, and an affine coupling layer. The splitting operation is excluded in the last red cube.}
  \label{fig:dia}
\end{figure*}

To incorporate spatial context into CD assessment,  Zhang and Wandell~\cite{zhang1997scielab} presented S-CIELAB, which extends CIELAB by adding spatial low-pass filtering as pre-processing. Similarly, Ouni~\etal~\cite{ouni2008new} provided a spatial extension of CIEDE2000.  Lee \etal~\cite{lee2005evaluation} re-examined histogram intersection, which is widely used in color image index, for the purpose of color image similarity assessment. Hong and Luo~\cite{hong2006new} chose to give larger weights to areas with spatially homogeneous colors and pixels with larger CDs. This method was later augmented by spatial filtering~\cite{pedersen2009new}. Lee and Plataniotis~\cite{lee2014towards} built upon the philosophy of color structural similarity, and gave the hue component careful treatment with circular statistics. Jaramillo~\etal~\cite{jaramillo2019evaluation} grouped the same texture areas for human-like CD assessment, using local binary patterns as texture descriptors.

General-purpose image quality models, including full-reference ones - SSIM~\cite{wang2004image}, VSI~\cite{zhang2014vsi},  LPIPS~\cite{zhang2018lpips} and DISTS~\cite{ding2020image}, reduced-reference ones - Wang05~\cite{wang2005reduced} and Yu09~\cite{yu2009method}, and no-reference ones - BRISQUE~\cite{mittal2012no}, NIQE~\cite{mittal2012making} and Gao13~\cite{gao2013no} can be directly adopted for CD assessment, regarding CDs as a particular form of ``visual degradations.'' Meanwhile, just-noticeable difference (JND) methods, \eg, Butteraugli~\cite{alakuijala2017guetzli} and \FLIP~\cite{andersson2020flip}, also attempt to characterize visually
indistinguishable color changes between two images. In the era of deep learning, due to the lack of sufficient human-labeled training data, DNN-based CD formulae are rarely proposed. Wang \etal~\cite{wang2022cdnet} created the first largest image dataset, SPCD, for perceptual CD assessment, and made one of the first attempts to train a DNN-based CD measure for photographic images. However, the underlying feature transform is not mathematically bijective.

\noindent\textbf{Normalizing Flow-based Models}. Normalizing
flow-based generative models are constructed by bijective functions $f: \mathbb{R}^D \rightarrow \mathbb{R}^D$, with typically easy-to-compute analytical inverse $f^{-1}: \mathbb{R}^D \rightarrow \mathbb{R}^D$. The primary goal of $f$ is to map raw data $\bm x$ to samples $\bm z = f(\bm x)$ from a simple probability distribution $p_\mathcal{Z}(\bm z)$. Many classic machine learning algorithms can be cast in  this framework, such as principal component analysis (PCA, where $f$ is a linear transform and $p_\mathcal{Z}(\bm z)$ is standard Gaussian) and independent component analysis (ICA, where $f$ is again linear and $p_\mathcal{Z}$ is factorized and heavy-tailed).

In 2014, Dinh \etal~\cite{dinh2014nice} proposed non-linear independent component estimation (NICE), as a generalization of ICA. NICE is considered the first normalizing flow with the introduction of the \textit{additive coupling} to ease the calculation of the Jacobian determinant. To make flow-based models more suitable for image-related tasks, Dinh \etal~\cite{dinh2017density} extended NICE to RealNVP, which admits a \textit{multi-scale autoregressive} architecture, implemented by \textit{squeezing} and \textit{affine coupling}. Kingma and Dhariwal~\cite{kingma2018glow} introduced the \textit{invertible $1\times1$ convolution} (\ie, the linear transform  in PCA) to replace the fixed random permutation for splitting the  channel dimension during multi-scale processing. The batch normalization in RealNVP is also replaced  with activation normalization (\ie, \textit{actnorm}). 
To allow unconstrained architectural design, Grathwohl~\etal~\cite{grathwohl2019ffjord} leveraged the Hutchinson's trace estimator for scalable and unbiased estimation of the log-density. Similarly, Behrmann \etal~\cite{behrmann2019invertible} proposed invertible residual networks (i-ResNet), introducing a tractable estimation to the Jacobian log-determinant of a residual block. Other representative normalizing flow work includes hierarchical recursive coupling~\cite{kruse2021hint} for increasing flow expressiveness, Wavelet Flow~\cite{yu2020wavelet} for scaling flow to ultra-high dimensional data, and Discrete Flow~\cite{tran2019discrete} for discrete data modeling. In this paper, we do not use normalizing flow for generative modeling, but for invertible feature transform.


\section{Proposed CD-Flow}
\label{sec:method}
In this section, we detail our proposed CD metric, CD-Flow, consisting of two key components: the feature transform and the distance measure~\cite{mahy1994cielab, luo2001ciede2000}.  The feature transform is built upon a variant of RealNVP~\cite{dinh2017density}, which is a learnable invertible transformation between input data and samples from a pre-fixed latent distribution. The CD distance is then computed using the Euclidean distance.

\subsection{Problem Definition}
We denote the RGB image space as $\mathcal{X}$ with an unknown distribution $p_\mathcal{X}$ and the transformed representation space as $\mathcal{Z}$ with a latent
distribution $p_\mathcal{Z}$. We are  
given a training dataset $\mathcal{D} = \{(\bm{x}^{(i)}, \bm{y}^{(i)}), \varDelta V^{(i)}\}_{i=1}^{M}$, where $\bm{x}^{(i)}, \bm{y}^{(i)} \in \mathcal{X}$ form the $i$-th image pair of the same visual content but different color appearances,  $\varDelta V^{(i)}$ represents the corresponding human perceptual CD collected from a subjective experiment, and $M$ is the number of training pairs. Our goal is to learn a normalizing flow-based invertible and differentiable transform $f$, which maps RGB images to latent representations with Gaussian conditionals for CD assessment.
\begin{figure}
  \centering
  \includegraphics[width=\linewidth]{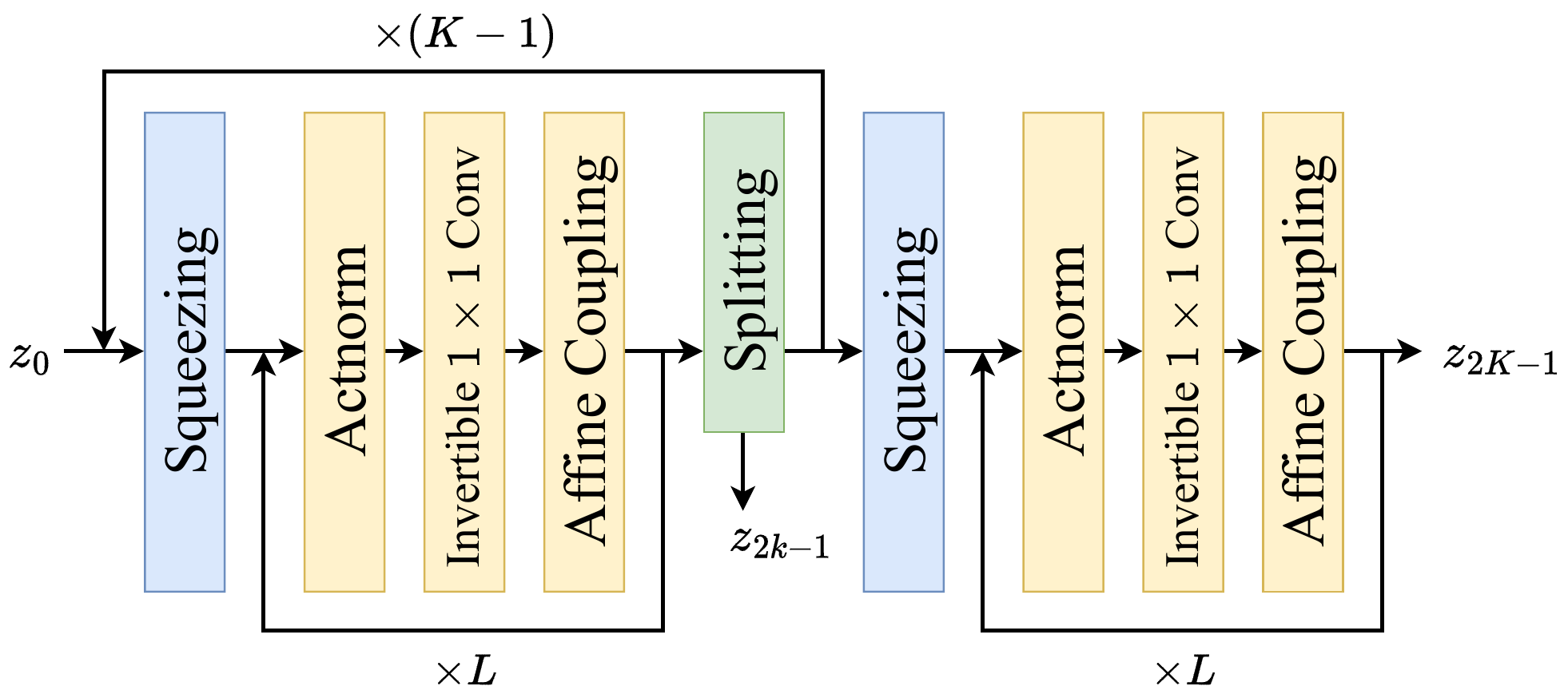}
  \caption{Architecture of the feature transform, adapted from~\cite{kingma2018glow}. The three yellow blocks constitute one step of flow.}
  \label{fig:tran}
\end{figure}

\subsection{CD-Flow}
\noindent\textbf{Feature Transform}.
Figure \ref{fig:dia} illustrates the system diagram  of the multi-scale autoregressive normalizing flow for the feature transform, which consists of $K$ scales of flow processing: $f=f_1\circ f_2\circ\cdots\circ f_K$ for multi-scale color and form interaction and abstraction. At the $k$-th scale, $\bm z_{2(k-1)}$ is processed and split into $\bm z_{2k-1}$ and $\bm z_{2k}$, the latter of which further undergoes the $k+1$-th scale of processing and splitting. At the final $K$-th scale, we only process $\bm z_{2(K-1)}$ to $\bm z_{2K-1}$ without splitting. By this notation, we assume that $\bm z_0 = \bm x$ is the input RGB image. The probability density of the latent representation $\bm z = \{\bm z_1, \bm z_3,\ldots, \bm z_{2K-1}\}$ can then be conditionally  factorized as 
\begin{align}
   p(\bm z) &= \prod_{k=1}^{K-1} p\left(\bm z_{2k-1}|\left\{\bm z_{\ge (2k+1)}\right\}\right)p(\bm z_{2K-1})\nonumber\\
   & = \prod_{k=1}^{K-1} p(\bm z_{2k-1}|\bm z_{2k})p(\bm z_{2K-1}).
\end{align}
The second equality is due to the bijectivity of the normalizing flow. $p(\bm z_{2k-1}|\bm z_{2k})$, for $k\in\{1,2,\cdots, K-1\}$, can conveniently be modeled as conditionally independent Gaussians. That is, the mean vector and the diagonal covariance matrix are computed from $\bm z_{2k}$ through say a tiny neural network~\cite{dinh2017density}. Similarly, $p(\bm z_{2K-1})$ is modeled as (unconditionally) independent Gaussians, where the parameters are directly estimated via backpropagation. As shown in Figure~\ref{fig:tran}, each scale of flow processing consists of a squeezing operation, multiple flow steps, and  a splitting operation.  Each flow step is further decomposed into three operations: \textit{actnorm}, \textit{invertible $1\times1$ convolution}, and \textit{affine coupling}.
\begin{itemize}
    \item \textbf{Actnorm} is introduced to replace the batch normalization to avoid degenerated performance when the mini-batch size is small. Given an input $\bm{z}$ of size $c\times h\times w$,
    the output of the same size can be computed by
    \begin{align}
    \bm{z}' &= \bm{s} \odot \bm{z} + \bm{t},
    \end{align}
with the  log-determinant $h\cdot w \cdot \mathrm{log}\vert\det(\mathrm{diag}(\bm{s}))\vert$.
    $\{\bm{s}, \bm{t}\}$ are learnable scaling and bias parameters. 
    \item \textbf{Invertible $1\times1$ Convolution} is a learnable  linear transform for channel mixing. Given a $c \times h\times w$ input $\bm{z}$ and a $c\times c$ weight matrix $\bm{W}$, we have
    \begin{align}
        \bm{z}' &= \bm{W}\bm{z},
    \end{align}
    with the log-determinant $h\cdot w\cdot\log\vert\mathrm{det}(\bm{W})\vert$.
    
    \item \textbf{Affine Coupling} constrains its Jacobian to be a triangular matrix, which facilitates the log-determinant computation. Specifically, the input $\bm{z}$ of $D$ dimensions is first split into two non-overlapping subsets: $\bm{z}_{1:d}$ and $\bm{z}_{d+1:D}$, where $d < D$. The output of the same dimension can be computed by
    \begin{equation}
    \begin{aligned}
        \bm{z}'_{1:d} &= \bm{z}_{1:d}, \\
        \bm{z}'_{d+1:D} &= \bm{z}_{d+1:D} \odot e^{s(\bm{z}_{1:d})} + t(\bm{z}_{1:d}),
    \end{aligned}
    \end{equation}
    where $s(\cdot)$ and $t(\cdot)$ are the scaling and translation functions, respectively, which are not necessarily invertible.
    The inverse is easily and analytically  derived as
    \begin{equation}
    \begin{aligned}
        \bm{z}_{1:d} &= \bm{z}'_{1:d}, \\
        \bm{z}_{d+1:D} &= \left(\bm{z}'_{d+1:D} - t(\bm{z}'_{1:d})\right) \odot e^{-s(\bm{z}'_{1:d})}.
    \end{aligned}
    \end{equation}
\end{itemize}

\noindent\textbf{CD Distance}.
We adopt the Euclidean distance, \ie, the square root of mean squared differences between two latent color representations $f(\bm{x})$ and $f(\bm{y})$, as the CD distance between two input images $\bm{x}$ and $\bm{y}$:\begin{align}\label{eq:exy}
    \varDelta E(\bm{x}, \bm{y}) = \sqrt{\frac{(f(\bm{x}) - f(\bm{y}))^T (f(\bm{x}) - f(\bm{y}))}{D}}.
\end{align}

\begin{table*}[t]
    \caption{STRESS, PLCC, and SRCC between predicted CDs ($\varDelta E$) and perceptual CDs ($\varDelta V$) in SPCD. The top section lists representative CD formulae developed from homogeneous color patches. The second section contains CD measures adapted for natural images. The third section includes general-purpose image quality models. The fourth section consists of JND measures. Closest to ours, CD-Net is a DNN-based data-driven CD measure for natural images. The top two methods are highlighted in boldface}
    \label{tab:comparision}
    \vspace{-.3cm}
	\begin{center}
	    \begin{threeparttable} 
		\begin{tabular}{l|ccc|ccc|ccc}
    		\toprule[1pt]
			\multirow{2}*{Method}  & \multicolumn{3}{c|}{Perfectly aligned pairs} & \multicolumn{3}{c|}{Non-perfectly aligned pairs} & \multicolumn{3}{c}{All}\\
    		\cline{2-10}
			  & STRESS$\downarrow$ & PLCC$\uparrow$ & SRCC$\uparrow$ & STRESS$\downarrow$ & PLCC$\uparrow$ & SRCC$\uparrow$ & STRESS$\downarrow$ & PLCC$\uparrow$ & SRCC$\uparrow$\\
           	\hline
 		    CIELAB~\cite{robertson1977cielab} &$31.244$ & $0.793$ & $0.775$ & $29.639$ &  $0.690$ & $0.579$  & $31.872$ & $0.716$ & $0.666$   \\
 		    CIE94~\cite{mcdonald1995cie94}  & $34.721$ & $0.790$ & $0.772$& $29.916$ & $0.693$ & $0.572$& $34.326$ & $0.710$ & $0.654$\\
                CIEDE2000~\cite{luo2001ciede2000}  & $29.975$ & $0.825$ & $0.821$& $30.347$ & $0.667$ & $0.563$& $31.439$ & $0.726$ & $0.686$\\	    			
 			\hline
 		    S-CIELAB~\cite{zhang1997scielab} &  $30.094$ & $0.822$ & $0.819$& $31.804$ &  $0.631$ & $0.522$ & $32.780$ & $0.700$ & $0.657$\\		    
 			Hong06~\cite{hong2006new} & $60.557$ & $0.794$ & $0.810$& $57.070$ & $0.543$ & $0.461$& $61.227$ & $0.645$ & $0.632$\\	
 			Ouni08\tnote{1}~\cite{ouni2008new} & $29.977$ & $0.826$ & $0.821$& $30.355$ & $0.668$ & $0.563$& $31.444$ & $0.726$ & $0.685$\\			
 		    Jaramillo19~\cite{jaramillo2019evaluation} &  $43.419$ & $0.514$ & $0.506$& $50.299$ & $0.081$ & $0.041$& $68.805$ & $0.321$ & $0.329$\\		    
 		    \hline
 		    SSIM~\cite{wang2004image} & $39.393$ & $0.589$ & $0.549$& $53.035$ & $0.077$ & $0.044$& $48.025$ & $0.309$ & $0.324$\\
 			\FLIP~\cite{andersson2020flip} & $29.318$ & $0.745$ & $0.715$ & $27.158$ & $0.734$ & $0.640$& $29.099$ & $0.718$ & $0.663$\\
 		    PieAPP~\cite{prashnani2018pieapp} & $29.044$ & $0.737$ & $0.737$ & $37.528$ & $0.522$ & $0.459$& $32.354$ & $0.652$ & $0.643$\\	    
 		    LPIPS~\cite{zhang2018lpips} & $44.811$ & $0.695$ & $0.688$ & $53.132$ & $0.219$ & $0.171$& $64.145$ & $0.439$ & $0.490$\\	    
 		    DISTS~\cite{ding2020image} & $31.409$ & $0.745$ & $0.746$ & $35.043$ & $0.528$ & $0.447$& $32.995$ & $0.640$ & $0.626$\\
 		    \hline
 			Chou07~\cite{chou2007fidelity} &  $50.721$ & $0.787$ & $0.785$& $36.184$ & $0.603$ & $0.459$& $49.545$ & $0.612$ & $0.557$\\
 		    Butteraugli~\cite{alakuijala2017guetzli} & $42.620$ & $0.606$ & $0.593$ & $48.217$ & $0.258$ & $0.245$& $54.737$ & $0.371$ & $0.359$\\
 		    \hline
			CD-Net~\cite{wang2022cdnet} &  $\mathbf{20.891}$  & $\mathbf{0.867}$ & $\mathbf{0.870}$ & $\mathbf{22.543}$ & $\mathbf{0.818}$ & $\mathbf{0.776}$& $\mathbf{21.431}$ & $\mathbf{0.846}$ & $\mathbf{0.842}$\\  
			CD-Flow &  $\mathbf{16.613}$  & $\mathbf{0.896}$ & $\mathbf{0.904}$ & $\mathbf{21.374}$ & $\mathbf{0.856}$ & $\mathbf{0.794}$& $\mathbf{18.473}$ & $\mathbf{0.871}$ & $\mathbf{0.865}$\\  
			\bottomrule[1pt]
		\end{tabular}
	    \begin{tablenotes}
            \item[1] The spatial extension of CIEDE2000.
        \end{tablenotes}
    \end{threeparttable}
	\end{center}
\end{table*}

\subsection{Loss Function}
It is straightforward to measure the $\ell_p$-norm induced distance between the predicted CD computed by Eq.~\eqref{eq:exy} and the perceptual CD of the given image pair $(\bm x, \bm y)$:
\begin{align}\label{eq:8}
    \ell(\bm x, \bm y) = \Vert\varDelta E(\bm x,\bm y) - \varDelta V(\bm x,\bm y)\Vert_p.
\end{align}
 To encourage the robustness of CD-Flow to mild geometric distortions, which are often unavoidable in practice, we introduce a multi-scale version of Eq.~\eqref{eq:8} to put more emphasis on coarser-scale latent representations:
\begin{align}\label{eq:9}
     \ell_{\mathrm{ms}}(\bm x, \bm y) = \sum_{k=1}^{K} \Vert\varDelta E_{k}(\bm x, \bm y) - \varDelta V(\bm x, \bm y)\Vert_p,
\end{align}
where 
\begin{align}
    \varDelta E_{k}(\bm x, \bm y) = \sqrt{\frac{(f_{k:}(\bm x) - f_{k:}(\bm y))^T (f_{k:}(\bm x) - f_{k:}(\bm y))}{D_{k}}}.
\end{align}
$f_{k:} (\bm x) = [\bm z_{2k-1}^T, \ldots, \bm z_{2K-1}^T]^T$, and $D_{k}$ is the number of feature dimensions of $f_{k:} (\bm x)$. That is, $\varDelta E_{k}(\bm{x},\bm{y})$ makes an estimate of $\varDelta V(\bm x,\bm y)$ using the $k$-th to $K$-th scale latent representations, and we have $\Delta E_1$ equal to $\Delta E$ in Eq.~\eqref{eq:exy}.

It is noteworthy that if the transform $f$ is supervised by the CD loss in Eq.~\eqref{eq:9} solely, the bijectivity of $f$ may not be necessarily guaranteed. This is because the adopted normalizing flow is not bijective by design. For example, the ``invertible '' $1\times 1$ convolution as the channel mixer becomes non-invertible when $\bm W$ is degenerate. Empirically, we observe that the scaling factor of the affine-coupling layer is close to zero after optimizing Eq.~\eqref{eq:9}. Since the inverse transform includes operations of dividing by the scaling factor, there will easily result in the exploding inverse problem~\cite{behrmann2021understanding}. Thus, we incorporate the commonly used maximum likelihood objective in normalizing flow~\cite{papamakarios2021normalizing} into our training objective. We work with the negative log-likelihood loss:

\begin{align}
\begin{split}
     \ell_{\mathrm{nl}}(\bm x) & = -\log p_{\mathcal{X}}(\bm x) \\
                  & = -\log p_{\mathcal{Z}}(f(\bm x))-\log \left|\text{det}\left(\frac{\partial f(\bm x)}{\partial \bm x}\right)\right|.
\end{split}
\end{align}
During training, we randomly sample a mini-batch $\mathcal{B}$ from the training dataset $\mathcal{D}$ in each iteration, and make use of a variant of stochastic gradient descent to optimize the parameters in CD-Flow:
\begin{align}
    \ell(\mathcal{B})&= \frac{1}{|\mathcal{B}|}\sum_{(x,y)\in \mathcal{B}}\bigg( \ell_\mathrm{ms}(\bm x,\bm y) + \lambda \big(\ell_{\mathrm{nl}}(\bm x) + \ell_{\mathrm{nl}}(\bm y)\big) \bigg),
\end{align}
where $|\mathcal{B}|$ denotes the cardinality of $\mathcal{B}$, and $\lambda$ is the trade-off to balance the magnitudes of different loss terms.

\begin{table*}[t]
    \caption{STRESS, PLCC, and SRCC between predicted CDs ($\varDelta E$) and perceptual CDs ($\varDelta V$) in SPCD~\cite{wang2022cdnet} under geometric distortions. Translation means randomly shifting one image with respect to the other image in an image pair  (by up to $5\%$ of pixels) in both spacial directions. Rotation means randomly rotating one image (by up to $3^{\circ}$) around the center point. Dilation means zooming in (and cropping) one image without changing the image size (by a factor of $1.05$)}
    \label{tab:geo}
    \vspace{-.3cm}
	\begin{center}
	    \begin{threeparttable} 
		\begin{tabular}{l|ccc|ccc|ccc}
    		\toprule[1pt]
			\multirow{2}*{Method}  & \multicolumn{3}{c|}{Translation} & \multicolumn{3}{c|}{Rotation} & \multicolumn{3}{c}{Dilation}\\
    		\cline{2-10}
			  & STRESS$\downarrow$ & PLCC$\uparrow$ & SRCC$\uparrow$ & STRESS$\downarrow$ & PLCC$\uparrow$ & SRCC$\uparrow$ & STRESS$\downarrow$ & PLCC$\uparrow$ & SRCC$\uparrow$\\
           	\hline
		    CIELAB\cite{robertson1977cielab} & $29.414$ & $0.620$ & $0.577$ & $32.633$ & $0.529$ & $0.495$ & $31.511$ & $0.519$ & $0.467$  \\  
            CIE94\cite{mcdonald1995cie94} & $29.141$ & $0.645$ & $0.596$ & $31.943$ & $0.566$ & $0.519$ & $30.323$ & $0.567$ & $0.505$   \\  
            CIEDE2000\cite{luo2001ciede2000} & $28.035$ & $0.654$ & $0.613$ & $31.255$ & $0.566$ & $0.527$ & $29.928$ & $0.566$ & $0.512$   \\  
            CD-Net\cite{wang2022cdnet} & $\mathbf{19.825}$ & $\mathbf{0.845}$ & $\mathbf{0.842}$ & $\mathbf{22.463}$ & $\mathbf{0.784}$ & $\mathbf{0.772}$ & $\mathbf{21.704}$ & $\mathbf{0.787}$ & $\mathbf{0.773}$\\
            \hline
            CD-Flow & $\mathbf{19.311}$ & $\mathbf{0.852}$ & $\mathbf{0.856}$ & $\mathbf{20.139}$ & $\mathbf{0.837}$ & $\mathbf{0.816}$ & $\mathbf{21.352}$ & $\mathbf{0.827}$ & $\mathbf{0.797}$\\  
			\bottomrule[1pt]
		\end{tabular}
    \end{threeparttable}
	\end{center}
\end{table*}

\section{Experiments}
In this section, we begin by presenting the experimental setups. We then compare the proposed CD-Flow with $15$ CD measures. We last conduct extensive ablation studies to justify the key designs of CD-Flow.

\subsection{Experimental Setups}

\noindent\textbf{CD Dataset}. 
We conduct experiments on SPCD \cite{wang2022cdnet}, so far the largest natural image dataset tailored for perceptual CD assessment. SPCD consists of $15, 335$ color images out of $1, 000$ distinct natural scenes spanning a variety of realistic shooting scenarios. A total of $30,000$ image pairs are sampled for human labeling, where $10,005$ are non-perfectly aligned image pairs. For each image pair, there are $20$ human ratings. We randomly split $70\%$, $10\%$, and $20\%$ of image pairs in SPCD as training, validation, and test sets, respectively, according to the image content to ensure content independence.

\noindent\textbf{Network Architecture}.
We employ a variant of RealNVP~\cite{kingma2018glow}, which consists of $K=6$ scales, and each scale includes a squeezing operation, $L=8$ steps of flow, and a splitting operation. We do not split at the last scale. The squeezing operation trades the spatial size for the channel number by transforming an $s\times s\times c$ tensor to an $\frac{s}{2}\times \frac{s}{2} \times 4c$ tensor, where $s$ is the spatial size and $c$ is the number of channels. 
The splitting operation divides the latent representation of the same scale into two halves along the channel dimension. One is used as the color features for CD assessment, and the other is subject to further processing. 

\noindent\textbf{Training and Testing Details}.
We employ the Adam as the stochastic optimizer, where the batch size is $4$, and the initial learning rate is $10^{-5}$ with a decay factor of $2$ for every $5$ epochs. We train CD-Flow for $50$ epochs in total.

\noindent\textbf{Evaluation Criteria}. We use three standard criteria to quantify the performance of CD-Flow against existing CD measures, including the standardized residual sum of squares (STRESS)~\cite{garcia2007measurement}, Pearson linear correlation coefficient (PLCC), and Spearman’s rank correlation coefficient (SRCC). STRESS measures both prediction accuracy and statistical significance, which is defined by
\begin{align}
            \mathrm{STRESS} = 100\sqrt{\frac{\sum_{i=1}^{M}(\varDelta E_i - F \varDelta V_i)^2}{F^2 \sum_{i=1}^{M}
        \varDelta V_i^2}},
\end{align}
where $M$ is the number of test pairs and $F$ is the scale correction factor between $\varDelta E$ and $\varDelta V$, defined as
\begin{align}
    F = \frac{\sum_{i=1}^{M}\varDelta E_i ^2}{\sum_{i=1}^{M} \varDelta E_i\varDelta V_i}.
\end{align}
STRESS generally ranges from $0$ to $100$, where a small value suggests a tight-fitting between model predictions and ground truths. SRCC and PLCC measure the prediction monotonicity and prediction linearity, respectively. Before calculating PLCC, we linearize model predictions by regressing a four-parameter monotonic function.

\subsection{Main Results}

\noindent\textbf{SPCD Results}.
We compare the proposed CD-Flow with $15$ existing CD measures which can be grouped into four categories: 1) CD measures for homogeneous color patches:
CIELAB~\cite{robertson1977cielab}, CIE94~\cite{mcdonald1995cie94} and CIEDE2000~\cite{luo2001ciede2000},
2) CD measures for natural images:
S-CIELAB~\cite{zhang1997scielab}, Hong06~\cite{hong2006new}, Ouni08~\cite{ouni2008new},  Jaramillo19~\cite{jaramillo2019evaluation} and CD-Net~\cite{wang2022cdnet},
3) full-reference image quality model:
SSIM~\cite{wang2004image}, \FLIP~\cite{andersson2020flip}, PieAPP\cite{prashnani2018pieapp}, LPIPS~\cite{zhang2018lpips} and DISTS~\cite{ding2020image}, and
4) JND methods:
Chou07~\cite{chou2007fidelity} and Butteraugli~\cite{alakuijala2017guetzli}. 
We employ official implementations of Butteraugli and \FLIP, and retrain  PieAPP, LPIPS, and DISTS on the same training set as CD-Flow. The rest methods are implemented and made publicly available by Jaramillo~\etal~\cite{jaramillo2019evaluation}. 

\begin{table*}[t]
    
    \caption{Generalizability evaluation of CD-Flow on the COM dataset and its four subsets: BFD-P, Leeds, Witt, and RIT-DuPont. PLCC on RIT-DuPont is indicated by ``---'' since it is not computable}
    \label{tab:cds}
    \vspace{-.3cm}
	\begin{center}
	    \begin{threeparttable} 
		\resizebox{\linewidth}{!}{\begin{tabular}{l|cccccccc|cc}
    		\toprule[1pt]
    		\multirow{2}*{Method} & \multicolumn{2}{c}{BFD-P \cite{luo1987bfd}} & \multicolumn{2}{c}{Leeds~\cite{kim1997leeds}} & \multicolumn{2}{c}{Witt~\cite{witt1999witt}} & \multicolumn{2}{c|}{RIT-DuPont~\cite{berns1991rit-dupont}} & \multicolumn{2}{c}{COM dataset~\cite{luo2001ciede2000}}\\
    		\cline{2-11}
			 & STRESS$\downarrow$ & PLCC$\uparrow$ & STRESS$\downarrow$ & PLCC$\uparrow$ & STRESS$\downarrow$ & PLCC$\uparrow$ & STRESS$\downarrow$ & PLCC$\uparrow$ & STRESS$\downarrow$ & PLCC$\uparrow$\\
           	\hline
			CIELAB \cite{robertson1977cielab} & $45.054$ & $0.749$ & $40.093$ & $0.295$ & $51.689$ & $0.565$ & $30.348$  & --- & $45.202$ & $0.693$ \\
			CIE94~\cite{mcdonald1995cie94} & $35.798$  & $0.830$ & $\mathbf{30.494}$  & $\mathbf{0.584}$ & $\mathbf{31.857}$  & $0.793$ & $\mathbf{20.982}$  & --- & $\mathbf{33.235}$& $\mathbf{0.814}$  \\
            CIEDE2000 \cite{luo2001ciede2000} & $\mathbf{31.935}$  & $\mathbf{0.861}$ & $\mathbf{19.247}$  & $\mathbf{0.772}$ & $\mathbf{30.358}$  & $\mathbf{0.825}$ & $\mathbf{20.239}$ & --- & $\mathbf{28.979}$ & $\mathbf{0.862}$ \\
			CD-Net & $39.312$ & $0.791$& $38.558$ & $0.449$ & $33.640$ & $\mathbf{0.828}$ & $42.999$ & --- & $38.872$& $0.786$ \\
                \hline
                CD-Flow & $\mathbf{34.661}$ & $\mathbf{0.833}$& $34.275$ & $0.476$ & $31.965$ & $0.820$ & $36.504$ & --- & $35.061$& $0.801$ \\
			\bottomrule[1pt]
		\end{tabular}}
    \end{threeparttable}
	\end{center} 
\end{table*}

The comparison results are documented in Table~\ref{tab:comparision}. Several interesting observations merit attention. First, CIE94 and CIEDE2000 rectify the non-uniformity of CIELAB on color patches by incorporating weight coefficients. Nonetheless, when it comes to natural photographic images, these approaches do not show noticeable improvements compared to the original CIELAB formula. This provides a strong indication that  humans perceive CDs of uniformly colored patches and natural images in drastically different manners. Na\"{i}ve extensions such as spatial filtering and texture grouping may not be sufficient to bridge the gap as evidenced by the results in the second section of Table~\ref{tab:comparision}. Second, despite being retrained on the same training set as CD-Flow, general-purpose DNN-based image quality models PieAPP \cite{prashnani2018pieapp}, LPIPS \cite{zhang2018lpips}, and DISTS \cite{ding2020image} do not deliver comparable performance. We believe this arises for PieAPP because it requires a much larger labeled set to support training from scratch, and $21,000$ training pairs, even with data augmentation, are less likely to overcome overfitting. LPIPS and DISTS rely on pre-trained VGG features, which are empirically proven to be more structure and texture inductive. As a result, they may be less ideal for assessing CDs.
Third, DNN-based CD formulae, \ie, CD-Net~\cite{wang2022cdnet} and CD-Flow, outperform all other competing models, which are primarily  attributed to their superior representation learning capabilities. This observation confirms the potential of employing DNNs in the field of color science.
Fourth,  CD-Flow achieves the most remarkable results, thereby demonstrating the promise of the multi-scale autoregressive normalizing flow for CD assessment.

\noindent\textbf{Robustness Results to Geometric Distortions}. We conduct experiments to evaluate the robustness of CD-Flow to mild geometric distortions, including translation, rotation, and dilation. Specifically, we translate randomly one image with respect to the other image in an image pair by up to $5\%$ pixels in both  horizontal and vertical directions, rotate randomly one image clockwise by up to $3^{\circ}$, and zoom in one image by a factor of $1.05$. The corresponding results are presented in Table~\ref{tab:geo}. We find that the performance of all models degrades due to the presence of geometric distortions. CIELAB-based formulae are particularly sensitive to geometric distortions because of pixel-by-pixel CD computation. In contrast, CD-Flow demonstrates the best robustness results, which is as expected because it is designed to  abstract color features, characterized by the interdependence between color and form for CD assessment.

\begin{table}[t]
    \caption{Generalizability evaluation of CD-Flow on the TID2013 subset, containing quantization noise, image color quantization with dither, and chromatic aberration artifacts}
    \label{tab:ds}
    \vspace{-.3cm}
	\begin{center}
	    \begin{threeparttable} 
     \small
		\begin{tabular}{l|ccc}
    		\toprule[1pt]
			Method  & STRESS$\downarrow$ & PLCC$\uparrow$ & SRCC$\uparrow$\\
            \hline
            CIEDE2000~\cite{luo2001ciede2000} & $18.203$ & $0.730$ & $0.751$\\
            PieAPP~\cite{prashnani2018pieapp} & $20.918$ & $0.620$ & $0.653$ \\
            LPIPS~\cite{zhang2018lpips} & $15.420$ & $0.816$ & $0.804$ \\
            DISTS~\cite{ding2020image} & $\mathbf{15.235}$ & $\mathbf{0.821}$ & $0.805$ \\
		CD-Net~\cite{wang2022cdnet} & $15.962$ & $0.801$ & $\mathbf{0.826}$\\
            \hline
            CD-Flow & $\mathbf{14.110}$ & $\mathbf{0.837}$ & $\mathbf{0.832}$\\
			\bottomrule[1pt]
		\end{tabular}
    \end{threeparttable}
	\end{center} 
\end{table}

\noindent\textbf{Generalizability Results on Other Datasets}. We examine the generalizability of CD-Flow on the COM dataset~\cite{luo2001ciede2000} and the TID2013 subset~\cite{Ponomarenko2015image}. The COM dataset consists of four subsets: BFD-P~\cite{luo1987bfd}, Leeds~\cite{kim1997leeds}, Witt~\cite{witt1999witt}, and RIT-DuPont~\cite{berns1991rit-dupont}, which include uniformly colored patches used for the development of CIEDE2000. The comparison results are reported in Table~\ref{tab:cds}. It is clear that two DNN-based CD models perform better than CIELAB on the COM dataset, despite not being exposed to color patch data during training. Although underperforming CIE94 and CIEDE2000, CD-Flow exhibits better generalization compared to CD-Net. We then test the generalizability of the TID2013 subset, which contains three types of color-related distortions: quantization noise, image color quantization with dither, and chromatic aberration. Table~\ref{tab:ds} lists the comparison results. We find that CD-Flow performs much better than the best-performing CIELAB-based CD formula CIEDE2000, and is on par with  general-purpose image quality methods LPIPS and DISTS, which may have been exposed to similar distortion appearances during training. All these results constitute supportive evidence that the multi-scale autoregressive normalizing flow provides a compelling embodiment of the hypothesis: ``color and form  are inextricably
interdependent as a unitray process of perceptual organization~\cite{shapley2011color,kanizsa1979organization}.''

\begin{figure*}
  \centering
  \includegraphics[width=\textwidth]{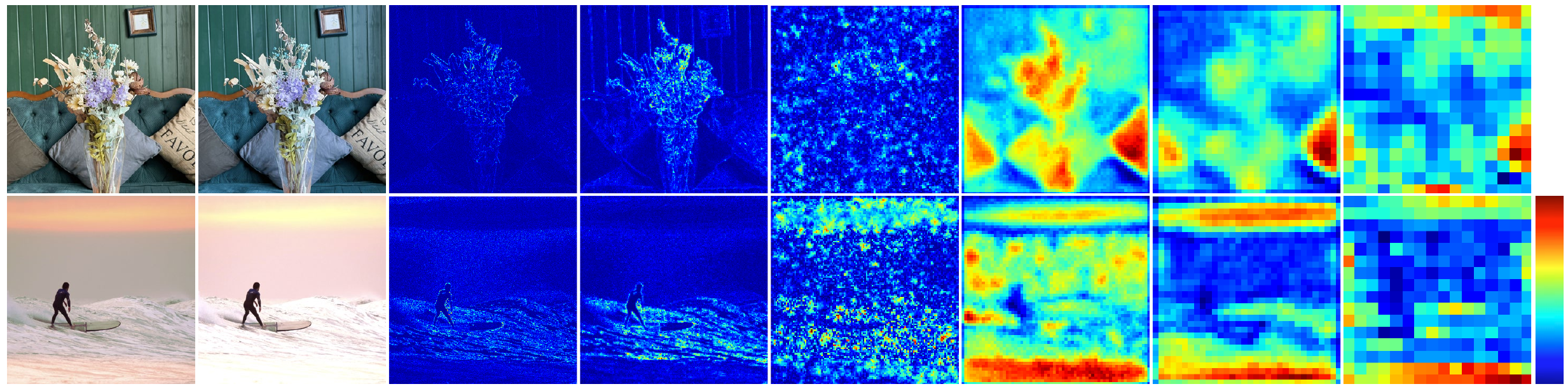}
  \caption{Multi-scale local CD maps generated by CD-Flow. First two columns: Input image pair $(\bm x,\bm y)$. Third to the eighth column: Six scales of CD maps generated by CD-Flow, with a warmer color indicating a larger local CD.}
  \label{fig:cdm}
\end{figure*}

\noindent\textbf{Visualization of Local CD Maps}. We further visualize the multi-scale local CD maps generated by CD-Flow, as illustrated in Figure~\ref{fig:cdm}.
Our observations are as follows. 
First, local CDs are generally small at the first three scales  compared to those at later scales, indicating that CD-Flow indeed relies more on coarser-scale representations to assess perceptual CDs, as encouraged by the multi-scale training objective in Eq.~\eqref{eq:9}. Second, interestingly, the effects of single-opponent and double-opponent cells emerge in the computation of CD-Flow. At coarser scales, local CDs are evaluated over a large image area, and exhibit reasonable sensitivity, resembling single-opponent cells in V1. At finer scales, local CDs are more responsive to textures and boundaries, consistent with the response characteristics of double-opponent cells in V1. 

\begin{table}[t]
    \caption{Ablation analysis of the number of flow steps in CD-Flow (\ie, the hyperparameter $L$), where the number of scales is fixed to six. The default setting is highlighted in boldface}
    \label{tab:ablation-step}
    \vspace{-.3cm}
	\begin{center}
	    \begin{threeparttable} 
		\begin{tabular}{l|ccc}
    		\toprule[1pt]
    		\# of flow steps & STRESS$\downarrow$ & PLCC$\uparrow$ & SRCC$\uparrow$\\
          	\hline
		    $L=2$  & $21.633$ & $0.837$ & $0.826$  \\  
            $L=4$ & $20.994$ & $0.838$ & $0.825$   \\  
            $L=\mathbf{8}$ & $18.473$ & $0.871$ & $0.865$   \\  
            $L=16$ & $17.792$ & $0.879$ & $0.870$ \\  
			\bottomrule[1pt]
		\end{tabular}
    \end{threeparttable}
	\end{center}
\end{table}

\begin{table}[t]
    \caption{Ablation analysis of the number of scales in CD-Flow (\ie, the hyperparameter $K$), where the number of flow steps is fixed to eight. The default setting is highlighted in boldface}
    \label{tab:ablation-scale}
    \vspace{-.3cm}
	\begin{center}
	    \begin{threeparttable} 
		\begin{tabular}{l|ccc}
    		\toprule[1pt]
			\# of scales & STRESS$\downarrow$ & PLCC$\uparrow$ & SRCC$\uparrow$ \\
           	\hline
		    $K=2$  & $24.686$ & $0.760$ & $0.732$ \\  
            $K=4$  & $19.730$ & $0.857$ & $0.850$ \\  
            $K=\mathbf{6}$  & $18.473$ & $0.871$ & $0.865$ \\  
            $K=8$  & $18.524$ & $0.874$ & $0.861$ \\  
			\bottomrule[1pt]
		\end{tabular}
    \end{threeparttable}
	\end{center}
\end{table}

\subsection{Ablation Studies}
We conduct two ablation experiments to evaluate the key design choices of CD-Flow. First, we study the impact of the number of flow steps (\ie, the hyperparameter $L$) on the model performance. In Table~\ref{tab:ablation-step}, we find that the  performance of CD-Flow increases with the number of flow steps, which in turn expands the non-linear representational capacity of CD-Flow. Nevertheless, too many steps may substantially increase the computational complexity, and negatively impact  generalization as well due to potential overfitting. We select $L=8$ as the default setting to strike a good balance between model performance and complexity.
Next, we examine the effect of the number of  scales (\ie, the hyperparameter $K$) on the model performance. Table~\ref{tab:ablation-scale} reports the results. We find that increasing the number of scales from $2$ to $6$ significantly improves the CD assessment performance. However, further increasing $K$ does not bring extra performance improvements, not as the case of increasing $L$. We attribute this to the setting of the current training input resolution (\ie, $768\times 768\times 3$). When $K=8$ and at the coarsest scale, we are essentially comparing the difference of two ``globally averaged'' color values over the entire images, which is less biologically plausible and less practically meaningful. 
\section{Conclusion}
We have introduced CD-Flow, a normalizing flow-based CD metric for photographic images. Our approach is inspired by the scientific evidence that humans perceive color based on the interdependence of color and form in images. We utilized a multi-scale autoregressive normalizing flow to learn a coordinate transform,  followed by computing the Euclidean distance in the transformed space. Remarkably, the learned feature transform enjoys four desirable properties: 1) consistent with the working mechanism of human color perception, 2)  proper as a mathematical metric, 3) accurate to explain human data of perceptual CDs, and 4) robust to slight geometric distortions.  We hope that the proposed CD-Flow can benefit related fields in imaging/illumination, vision science, and color science.

\section*{Acknowledgement}
We would like to thank Keshuo Xu for the technical support and code sharing. This work was supported in part by the National Natural Science Foundation of China under Grant
62071407, the Hong Kong  ITC Innovation and Technology Fund (9440288), and the CUHKSZ University Development Fund (UDF01002437).

{\small
\bibliographystyle{ieee_fullname}
\bibliography{egbib}
}

\end{document}